\documentclass[10pt,twocolumn,letterpaper]{article}
\usepackage[margin=0.8in]{geometry}
\usepackage{times}
\usepackage{epsfig}
\usepackage{graphicx}
\usepackage{amsmath}
\usepackage{amssymb}
\usepackage[ruled]{algorithm2e}
\usepackage{tabulary}
\usepackage{threeparttable}  
\usepackage[super,comma,sort&compress,numbers]{natbib}
\usepackage{subcaption}
\usepackage{pifont}
\usepackage{titlesec}
\usepackage{mathpazo}
\usepackage{helvet}
\usepackage[labelfont=bf,textfont=it]{caption}

\usepackage{color}
\usepackage[table,xcdraw]{xcolor}
\definecolor{grey}{rgb}{0.8,0.8,0.8}
\newcommand{\cmark}{\ding{51}}%
\newcommand{\xmark}{\ding{55}}%
\graphicspath{{Figures/}}
\titlespacing\section{0pt}{6pt plus 2pt minus 0pt}{6pt plus 1pt minus 0pt}
\titlespacing\subsection{0pt}{6pt plus 2pt minus 0pt}{6pt plus 1pt minus 0pt}
\titlespacing\subsubsection{0pt}{3pt plus 0pt minus 0pt}{2pt plus 0pt minus 0pt}
\titleformat{\section}{\large\bfseries\sffamily}{\thesection}{1em}{}
\titleformat{\subsection}{\normalsize\bfseries\sffamily}{\thesubsection}{1em}{}
\titleformat{\subsubsection}{\small\sffamily}{\thesubsubsection}{1em}{}

\begin{document}

\newcommand{\comment}[1]{}
\title{Multi-level 3D CNN for Learning Multi-scale Spatial Features}


\author{\small Sambit Ghadai\\
\and
\small Xian Yeow Lee\\
\and
\small Aditya Balu\\
\and
\small Soumik Sarkar\\
\and
\small Adarsh Krishnamurthy\\
\and
{\tt\small [sambitg | xylee | baditya | soumiks | adarsh]@iastate.edu}\\
\small Iowa State University, 
\small Ames, IA\\
}
\date{ }

\maketitle

\thispagestyle{empty}


\begin{abstract}
3D object recognition accuracy can be improved by learning the multi-scale spatial features from 3D spatial geometric representations of objects such as point clouds, 3D models, surfaces, and RGB-D data. Current deep learning approaches learn such features either using structured data representations (voxel grids and octrees) or from unstructured representations (graphs and point clouds). Learning features from such structured representations is limited by the restriction on resolution and tree depth while unstructured representations creates a challenge due to non-uniformity among data samples. 
In this paper, we propose an end-to-end multi-level learning approach on a multi-level voxel grid to overcome these drawbacks.
To demonstrate the utility of the proposed multi-level learning, we use a multi-level voxel representation of 3D objects to perform object recognition. The multi-level voxel representation consists of a coarse voxel grid that contains volumetric information of the 3D object. In addition, each voxel in the coarse grid that contains a portion of the object boundary is subdivided into multiple fine-level voxel grids. The performance of our multi-level learning algorithm for object recognition is comparable to dense voxel representations while using significantly lower memory.
\end{abstract}
\section{Introduction}

\comment{
1. Spatial data learning is good and it is different from Multi-View Learning and graph based learning
2. How hierarchical learning is important in spatial data. mention about memory, density/non-uniformity of features, scalability, varying bounding box, structure in unstructured at the same time unstructured in structured. hierarchy is useful... but too much hierarchy is further difficult. flexibility is important
3. Explain about multi-level voxel learning, show comparison with other methods. Establish that the only major comparison should be OctNet
4. Need for interpretable/attention based feedback for learning for creating human interpretable and sane ML models.
5. Summary of all the key contributions
6. Flow of the rest of the paper
}

A three dimensional object comprises of a different multi-scale features inherent to its geometry and its overall shape. Deep Neural Networks have been used to extract meaningful information from spatial data and perform object recognition. Several works have made substantial efforts to perform object recognition from 3D data by extending image recognition principles such as projection of the 3D information to 2D or 2.5D (depth inclusion) images~\cite{wu20153d,su15mvcnn} and multiple 2D views of the 3D object~\cite{li2018sonet, kanezaki2016rotationnet, SFIKAS2018208, qi2016volumetric}. Though this is effective in many applications including 3D reconstruction, some spatial relationships among the features get lost and this makes it infeasible for certain problems such as graphics rendering~\cite{tatarchenko2017octree}, point cloud labeling~\cite{Qi2017PointNetDH}, design and manufacturing~\cite{GHADAI2018263}. However, a major limitation in learning directly from 3D data is the high memory requirement. The presence of abundant information in spatial data coupled with the large data requirement for efficient training of deep learning algorithms render this task impractical for high-resolution 3D data.


\begin{figure*}[ht]
 \centering
 \includegraphics[width=0.9\textwidth,clip,trim={0in 0.5in 0in 2.5in}]{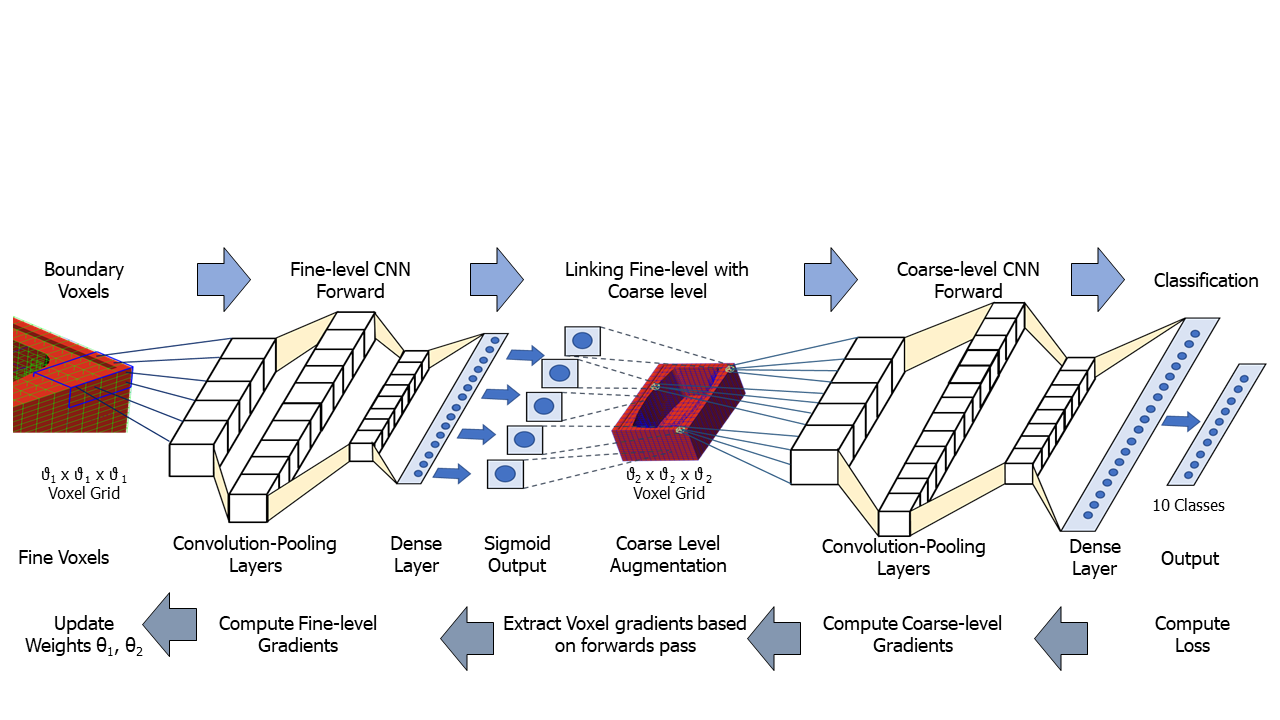}
 \caption{Multi-Resolution Convolutional Neural Network (MRCNN). Our proposed network can learn from a hierarchical data representation with a coarse level of information and information of boundary voxels which connects to the information from fine level voxels. For a forward pass (left to right) the information learnt from selected fine level voxels using the \emph{fine-level} CNN is embedded in the coarse level input to \emph{coarse-level} CNN and then the final prediction is obtained. The backward pass follows the reverse order of the forward pass (right to left).}
 \label{Fig:MultiLevelCNN}
\end{figure*}

Convolutional Neural Networks (CNNs) are natural candidates for this task as they have been proven to be effective for learning features from 3D spatial data~\cite{ioannidou2017deep,7353481}. However, training CNNs using uniform data representations (such as voxels) become inefficient when spatial features exist on different physical scales since uniform data representation cannot effectively accommodate this non-uniformity~\cite{spatialbook}. Hence, efficient and scalable deep learning techniques that exploit sparse and hierarchical data representations are necessary to deal with large 3D data sets. The most common high resolution voxel representation of 3D geometries is Octree~\cite{meagher1982geometric}, which is a structured representation that recursively divides each voxel into 8 sub-voxels and stores them in a tree structure. Octree based learning frameworks like OctNet~\cite{riegler2016octnet} require custom convolution operations specific for the octree data structure. This approach facilitates learning from high-resolution structured data.

In this paper, we present a novel approach to enable hierarchical learning of features from a flexible multi-level unstructured voxel representation of spatial data. We achieve this by adopting the multi-level voxelization framework developed by Young et. al~\cite{gavin2018cg}. A multi-level voxel grid is defined as a binary occupancy grid at two levels to represent a 3D object with two independent user-defined resolutions of voxel grids. We developed a multi-level CNN that can effectively learn features despite the unstructured nature of the multi-level data representation. 

\comment{
The specific contributions of this paper include:
\begin{enumerate}
\setlength\itemsep{0.0em}
\item A framework for learning from hierarchical multi-scale representation of 3D spatial data, where there are two levels of information description, a coarse level and a fine level. The coarse level description of features in the spatial data can give an overall understanding of the sample, while some important finer level features could improve the learning by augmenting the coarse level description.
\item A new method of connecting the multiple learning levels via salient information flow between them. 
We develop the connection between the two levels of the network using feedback from the coarse level to the fine level provided by interpretability mechanisms such as GradCAM (Gradient of the class activation map) regarding important spatial regions.
\item Demonstration of the above mentioned methods on a multi-level voxel representation used for the purpose of object recognition of ModelNet10 and ModelNet40 datasets. We achieve similar accuracy of results as previously reported, while using considerably lower memory during training.
\end{enumerate}

The paper is arranged as follows. In Section~\ref{Sec:RelatedWork}, we discuss a few relevant works in the field of learning from 3D data using different data representations. In Section~\ref{Sec:Voxelization}, we describe the multi-resolution representation of 3D data using multi-level voxelization, which we generate using a GPU accelerated voxelization algorithm. We explain the Multi-Resolution Convolutional Neural Network (MRCNN) architecture, that effectively learns the features from the multi-level voxel representation, in Section~\ref{Sec:MRCNN}. Finally, in Section~\ref{Sec:Results}, we present the results from evaluating the MRCNN on multi-level voxel data to classify objects in the ModelNet10 dataset and also explain the effectiveness of MRCNN to learn from 3D data with reduced memory usage.

\section{Related works}
\label{Sec:RelatedWork}

\begin{table*}[!t]
\caption{Comparisons between different spatial deep learning approaches. Our approach (MRCNN) retains some of the advantages OctNet of such as memory efficiency while still having a flexible data structure. MRCNN also enables spatial multi-scale learning from multi-resolution data.}
\begin{center}
\resizebox{\linewidth}{!}{
\tiny
\begin{threeparttable}
\label{table1}
  \begin{tabular}{l l l c c}
    \hline 
    Method & Data Representation & Data Structure & Memory Efficient & Spatial Multi-scale Learning \\ \hline
    VoxNet & Voxels & Structured & \xmark & \xmark \\
    PointNet++ & Point cloud & Unstructured & \xmark & \xmark  \\
    OctNet & Octree & Structured & \cmark & \xmark  \\
    \textbf{MRCNN} & Multi-level voxels & Structured-flexible & \cmark & \cmark \\
    \hline
  \end{tabular}
\end{threeparttable}
}
\end{center}
\end{table*}

Learning from spatial data has been an active research topic and several approaches have been developed by researchers to address this challenge. Most of the approaches can be categorized into two main learning methodologies. The first category of approaches are based on learning from unstructured spatial data such as point clouds~\cite{8099499, Qi2017PointNetDH, DBLP:journals/corr/KlokovL17, Rov18b}, meshes~\cite{Fey2017SplineCNNFG, Masci2015GeodesicCN, Boscaini:2016:LSC:3157382.3157455} and graphs~\cite{7974879, doi:10.1111/cgf.12064, Wang2018LocalSG}. The second class of methods use structured spatial data such as voxel grids~\cite{wu20153d, 7353481, 8403889, 7900177, Wu:2016:LPL:3157096.3157106}, octrees~\cite{riegler2016octnet, tatarchenko2017octree, Wang-2017-ocnn}, RGB-D images~\cite{Socher:2012:CDL:2999134.2999208, Couprie2013IndoorSS}, etc. Using general deep learning methods (such as CNNs) to learn from unstructured spatial data is challenging, since many of the operations require a structured input. Our work is focused on a sub-class of structured methods that use volumetric representation (voxels, octrees) to learn from spatial data, while also being flexible in terms of the data structure. Our multi-level voxel data structure makes use of user-defined voxel resolution at each level, making it more flexible than the octree data structure (each voxel is divided exactly into $2^3$ sub-voxels). This allows us to achieve very high effective resolutions using only two levels, while retaining the memory efficiency.

\subsection{Multi-level voxel learning}
Learning from 3D voxel data was initially explored by Wu et. al (3DShapeNet) using convolutional deep belief networks (CDBNs)~\cite{wu20153d} and later by Maturana et. al (VoxNet)~\cite{7353481} using 3D-CNN. It is challenging to achieve a  good classification accuracy on the ModelNet10 and ModelNet40 datasets using these approaches. Most approaches that use voxel data use a maximum of $32^3$ resolution; recently with the increase in GPU memory, using a resolution of $64^3$ is possible. However, increasing the resolution beyond that is not practical with current systems for training deep networks using dense voxel grids. 

Increasing the effective voxel resolution beyond $64^3$ require new specialized data structures. Riegler et. al developed OctNets~\cite{tatarchenko2017octree,riegler2016octnet}, which make use of an octree-based voxel representation of the data. In a tree structure each voxel is represented as a node and each of those nodes are connected to exactly eight subdivided voxels or octants. However, traversing the octree structure requires recursive algorithms. OctNets solved this problem by making use of a shallow octree data stored using a regular grid. This allowed them to directly index the data in the octree without recursively parsing the tree and were able to achieve an effective resolution of $256^3$. In this work, we make use of a different strategy of having multiple levels of voxelization that can have arbitrary resolutions at each level. Using this approach, we can also achieve very high resolution similar to those achieved using OctNets. For example, we can achieve effective resolution of $256^3$ by having a coarse resolution of $64^3$ and a fine resolution of $4^3$.

\subsection{Feedback from interpretable attention maps}

Convolutional Neural Networks (CNNs)~\cite{726791, Krizhevsky:2017:ICD:3098997.3065386, He2016DeepRL, Huang2017DenselyCC} are the most preferred vision detectors when it comes to deep learning based computer vision such as image classification, segmentation and detection. However, CNNs are considered to be black boxes since they do not provide insights into the workings of the feature extractors and the hidden layers. 
To overcome this setback, a plethora of work~\cite{Zhang2018} have been dedicated in recent times to unlock this black box nature of CNNs for more explainable/interpretable insights. Most of the interpretability methods~\cite{Zeiler-2014, olah2017feature, 8237336, Ribeiro-2016, Zhang2018ExaminingCR} are based on understanding a pre-trained network; although, a few works present disentangled/interpretable representations of neural networks that produce self-explanatory graphs. Using these disentangled representations, researchers have come up with middle-to-end learning models that use weak to strong supervisory inputs from interpretable models to enhance learning. These supervisory inputs are in the form of feedback cues, such as active question-answering and object annotations, to enable hierarchical object understanding~\cite{8099897}, one-shot/multi-shot learning~\cite{Zhang2017InteractivelyTC}, etc. In this paper, we show the use of such an interpretability mechanism to augment 3D-CNNs with a feedback based on attention maps. In accordance to the multi-level data representation we use in our work, this feedback mechanism is used to bridge the fine-level learning to the coarse-level learning.

}

\section{Multi-level Voxelization}
\label{Sec:Voxelization}

\begin{figure*}[t!]
 \centering
 \begin{subfigure}[b]{0.21\textwidth}\centering
  \includegraphics[width=1.0\textwidth, trim = 0.2in 0.2in 0.2in 0.2in, clip]{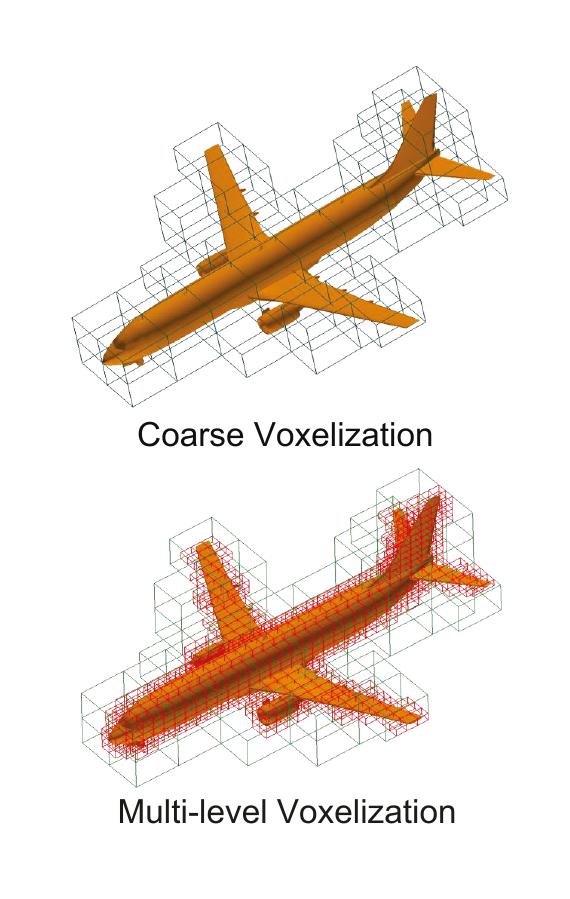}
  \caption{}
 \end{subfigure}
 \hfill
 \begin{subfigure}[b]{0.74\textwidth}\centering
  \includegraphics[width=1.0\textwidth, trim = 0in 0in 0in 0in, clip]{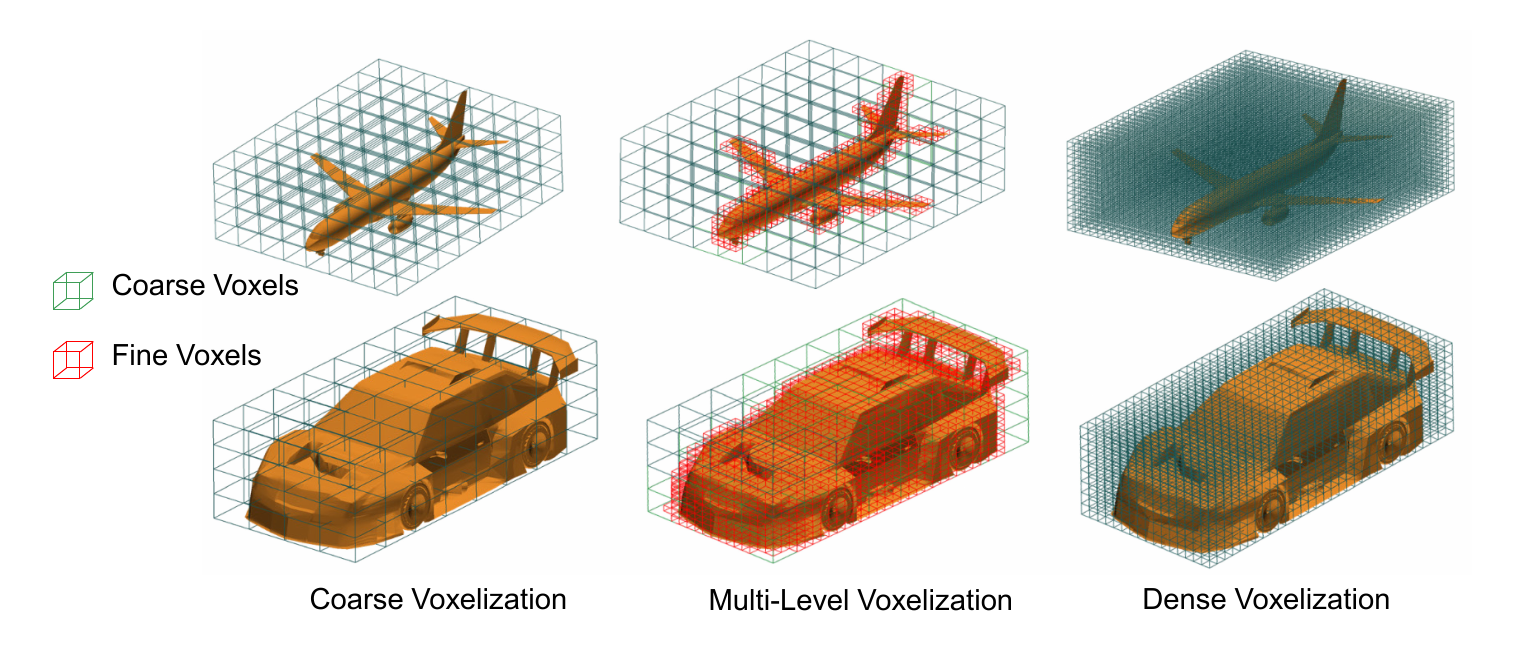}
  \caption{}
 \end{subfigure}
 \caption{Multi-level voxelization of B-rep CAD models. The fine level voxelization is performed only near the boundaries of the coarse level voxelization. The final resolution is equivalent to having dense level voxels throughout the model.}
 \label{Fig:MultiLevelvox}
\end{figure*}

In this section, we briefly describe the GPU-accelerated algorithm~\cite{gavin2018cg} we used to generate the multi-level voxelization from boundary representation(B-rep) of a 3D model. The multi-level voxelization is a binary occupancy grid having two major components namely, \emph{coarse-level voxelization} and \emph{fine-level voxelization}. The \emph{coarse-level} voxel grid represents the whole 3D CAD model at a coarse resolution and the \emph{fine-level} voxel grid represents the boundary of the \emph{coarse-level} voxel grid at a finer resolution in a hierarchical manner. The two levels of voxel grids are mapped to each other using a prefix-sum array mapping. For example, a CAD model can be represented at the \emph{coarse-level} with a voxel resolution of $32\times32\times32$ and each of the coarse boundary voxels can be further voxelized at a resolution of $4\times4\times4$ (see Figure~\ref{Fig:MultiLevelvox}). This makes the CAD model to be represented with an effective resolution of $128\times128\times128$ using the multi-level voxelization. We use a multi-level voxel data structure to store information pertaining to the geometry of an object in two hierarchical levels, thus exploiting the sparse nature of the data.

\comment{
Multi-resolution voxel representations of B-rep CAD models are shown in Figure~\ref{Fig:MultiLevelvox} along with corresponding coarse and dense voxel grids. At each level, the voxel occupancy is represented using a binary value of $0/1$ that defines whether a voxel is occupied by the object or not. To create the coarse-level voxelization, a standard grid of voxels of an user-defined resolution is constructed in the region occupied by the object (denoted by the axis-aligned bounding box (AABB)). A \emph{triangle-box} intersection test is used in the next step to identify the boundary voxels by checking the intersection of each triangle of the B-rep model with every voxel. Identifying the boundary voxels enable further division of the coarse-level grid into a fine resolution only at the boundary of the object without cubically increasing the voxel count. Once the boundary voxels are identified, they are further sub-divided to construct the fine-level voxel grid only at the boundary locations. The same \emph{triangle-box} intersection test is then used to identify the fine-level voxels that intersect with the triangles of the B-rep model. Further, an index array is created using \emph{exclusive prefix sum} that maps the memory location of each coarse-level boundary voxel to its corresponding fine-level voxel grid. The complete voxelization framework is GPU-accelerated, where each \emph{triangle-box} intersection test is computed in parallel at individual levels. Using this method on the GPU, a high-resolution (effective resolution of $128^3$) multi-level voxelization of a low-mid polygon count model can be generated in $180ms$.

Representing a 3D model using a multi-level voxel grid structure exploits the sparse nature of the data. When spatial data is represented using voxel grids, the voxel count increases as $\mathcal{O}(n^3)$ with increase in resolution of the grid and in turn reduces the total voxel space occupied by the object. To prove this, we representing the coarse-level voxel grid by $G_c$ and its corresponding fine-level voxel grid by $G_f$ with resolutions of $n_c\times n_c\times n_c$ and $n_f\times n_f\times n_f$, respectively. Let the number of boundary voxels be $\phi_b$. Comparing it with a dense voxel grid of resolution $n_d\times n_d \times n_d$ where $n_d = n_c\times n_f$, we see that, total voxels in multi-level data structure is $\phi_b\times n_d^3$. As the number of boundary voxels can never be more than the total voxel count, $1\leq \phi_b \leq n_c^3$. Hence,
\begin{align*}
    \phi_b\times n_d^3 &\leq n_c^3\times n_d^3 \\
    \Rightarrow\text{Voxel count}_{\text{multi}} &\leq \text{Voxel count}_{\text{dense}}
\end{align*}
}
\section{Multi-resolution CNN}
\label{Sec:MRCNN}


The multi-resolution convolutional neural network (MRCNN) consists of two 3DCNNs, with each CNN kernels performing 3D convolution operations, to learn the features in each level of the voxel grid. One of these 3DCNNs, named as \emph{Coarse-level CNN}, takes in the coarse level voxels as input while the other 3DCNN called \emph{Fine-level CNN} takes the fine level voxels as input. These two neural networks are intelligently combined to work together as a single unit in both forward pass and backward pass of the algorithm. This facilitates optimal learning from a multi-level data representation. 

The forward computation of MRCNN starts by learning from the fine-level voxel grids by randomly sampling a subset, $ \phi$, of the total boundary voxels, $\Phi$, in a 3D voxelized model. Each of these $\phi$ boundary voxels, with individual fine voxel grid $\vartheta_{2}$, are used as input to \emph{Fine-level CNN}. The \emph{Fine-level CNN} consists of blocks of \emph{convolution - max pooling} layer pairs and \emph{fully connected} layers connected sequentially, each with a ReLU activation function associated with it. \emph{Fine-Level CNN} outputs a single real numbered value $\eta_b$ for each of the selected boundary voxels $\Phi$. We replace the original coarse voxel grid values with $\eta_b$ at the corresponding voxel positions. This is performed with the help of the prefix sum based index arrays of the multi-level voxel grid as explained in~\cite{gavin2018cg}.

In the next phase of the MRCNN forward computation, the coarse-level voxel grid with selective embedding of the fine level voxel information $\eta_b$, is used as an input to the \emph{Coarse-level CNN}. The architecture of \emph{Coarse-level CNN} network comprises of different set of \emph{convolution - max pooling} layers. The end of the network has multiple \emph{fully connected} layers and the output is the class prediction probability vector. Categorical cross-entropy loss function is used to compute the loss of between predicted classes and true class labels. The forward pass of MRCNN network algorithm is illustrated in Figure~\ref{Fig:MultiLevelCNN}.
\comment{
\begin{algorithm}
    \textbf{Forward Pass:}\\
    \Indp
	\ForAll{Boundary voxels, $\phi_b$ \emph{\textbf{in parallel}}}{
				$\eta_b $= {$Forward_{Fine-level CNN}$($\vartheta_{2_{b}})$)};\\
                $\vartheta_1(\phi_b) = \eta_b$
	}
    $y_{pred} = Forward_{Coarse-level CNN}(\vartheta_1)$\\
    \Indm
    \textbf{Backward Pass:}\\
    \Indp
    $d\vartheta_1 = Backward_{Coarse-level CNN}(\vartheta_1, dy_{pred})$
	
	\ForAll{Boundary voxels, $\phi_b$ \emph{\textbf{in parallel}}}{
				$d\eta_b = d\vartheta_1(\phi_b)$\\
                $d\vartheta_{2_{b}} = Backward_{Coarse-level CNN}(\vartheta_{2_{b}}, d\eta_b)$;\\
                
	}
    \Indm
	\caption{$\text{MRCNN}$}
	\label{Alg:MRCNN}
\end{algorithm}
}

Once the forward computation of the MRCNN is established, the only challenge is to link the two networks such that the gradients can passed on from the coarse level network to the fine level network during back-propagation. This link is essential for obtaining gradients for the weights of the fine level network. The final loss between the $y_{pred}$  and $y_{true}$ of the coarse level network is first computed using categorical cross-entropy loss. Back-propagating this loss through the coarse level network is trivial. 
Once we obtain the gradients for input coarse level voxel embedding, we compute the gradient of $\eta_b$ and use that to backpropagate the same in the fine level voxel grid. Let the gradient of the loss with respect to coarse input be $d\theta_1$, using prefix sum, we track the gradients of the outputs of fine level network ($\eta_b$) and use it to back-propagate through the network. 


It is also worthwhile to note that since the same \emph{Fine-level CNN} is shared among all the boundary voxels, the gradients of $\theta_2$ for \emph{Fine-level CNN} are computed for all boundary voxels only once.

With the gradients linked, the network could be trained end-to-end to update its weights $\theta_1$ and  $\theta_2$ in such a way that the loss $L$, of the final prediction is minimized. 
The network parameters' update could be performed using the \emph{Adam} optimizer~\cite{kingma2014adam}. The complete operation of MRCNN is explained schematically in Figure~\ref{Fig:MultiLevelCNN}.

\comment{
\begin{figure}[!ht]
 \centering
 \includegraphics[width=0.45\textwidth,clip,trim={2.5in 1.0in 2.5in 1.0in}]{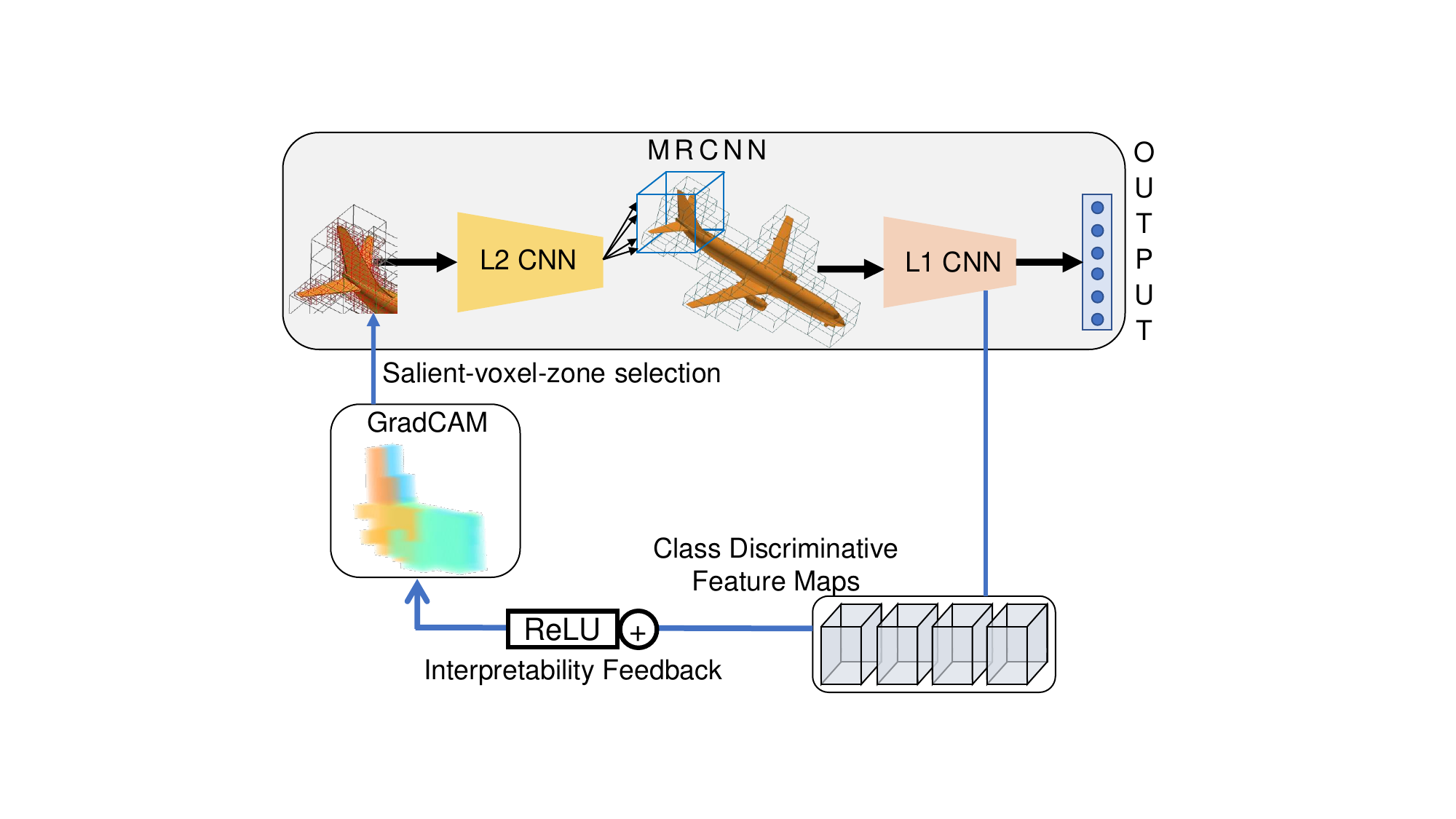}
 \caption{Interpretability based feedback for Multi-Resolution CNN. The output is first evaluated with just the coarse voxels using L1 CNN. The interpretability of the L1 CNN provides feedback salient features to select voxel-zones (SVZs) to train MRCNN as explained in  Algorithm~\ref{Alg:Feedback}.}
 \label{Fig:feedback}
\end{figure}

\subsection{Adaptive salient-voxel-zone feedback via interpretable attention maps}
In the forward computation of MRCNN as discussed in Section~\ref{Sec:forward}, we randomly sample $\phi$ boundary voxels from coarse-level voxel grid as the input to the \emph{Fine-level CNN} initially. This is done so as not to destroy the information present in the coarse-level voxel grid. In addition, we found that using all the $\phi$ voxels did not improve the learning capability of MRCNN substantially. However, the $\phi$ voxels sampled randomly might not be the perfect candidates that carry important spatial information in the fine-level grid. Hence, we propose an adaptive \emph{salient-voxel-zone} (SVZ) selection technique that samples the coarse-level boundary voxels based on an interpretability algorithm. These $\phi_{int}$ voxels, that are a fraction of the total $\Phi$ boundary voxels, are input to the \emph{Fine-level CNN}; with the rest of the forward and backward computations of MRCNN similar to the algorithms discussed in Agorithm~\ref{Alg:MRCNN}, respectively. In this section, we explain the process of using interpretability as a feedback to MRCNN for intelligent sampling of boundary voxels from the coarse-level grid.

The interpretability mechanism adopted to detect important boundary voxels is an enhancement of GradCAM~\cite{8237336} for 3D objects (3DGradCAM~\cite{GHADAI2018263}). 3DGradCAM backpropagates the gradients of class specific feature maps, retrieved from the CNN, with respect to the loss prediction to generate class activation maps (CAM) that estimate the regions in the input 3D object. The feature maps are first obtained from the CNNs and then a class discriminative gradient is computed with respect to these feature maps. Through this, spatially important feature maps are obtained for a particular class. Mapping the CAM to the input voxel grid allows us to extract regions, which we call \emph{salient-voxel-zones}, from the coarse-voxel grid along the boundary voxels.

\begin{algorithm}[t!]
    $\theta_1 \leftarrow Pretrain_{Coarse-level CNN}$\\
    \ForAll {Epochs}{
    $\gamma = 3DGradCAM(\vartheta_1(\Phi), \theta_1)$\\
    $y_{pred} \leftarrow Forward_{MRCNN}(\vartheta_2(\gamma(\phi_b)),\vartheta_1)$\\
    $L = Cross-entropy(y_{true}, y_{pred})$\\
    $\theta_1, \theta_2(Update) \leftarrow Backward_{MRCNN}(Loss)$\\
	}
	\caption{$\text{Feedback}_{\text{MRCNN}}$}
	\label{Alg:Feedback}
\end{algorithm}

In the first step to extract SVZ, we use a pre-trained model of \emph{Coarse-level CNN}, trained on coarse level voxel grid $\vartheta_1$ to generate the CAMs, $\gamma$, using 3DGradCAM. To overlay the CAM to the input data, we interpolate it to match the coarse level voxel grid size. This acts as an attention map of the voxel grid to interpret the important voxels contributing to the classification. The relevant voxels interpreted from $\gamma$ have a higher activation than the non-relevant voxels. We exploit this factor to choose $\gamma(\phi_b)$ voxels from the total boundary voxels $\Phi$ to be used as input to \emph{Fine-level CNN}. Then the forward computation of MRCNN proceeds as discussed in Section~\ref{Sec:forward}, to classify the object using the network parameters $\theta_1, \theta_2$. After computing the classification loss $L$, the backward computation of MRCNN updates the network parameters $\theta_1, \theta_2$ by back-propagating $L$ to the input layer of \emph{Fine-level CNN}. This constitutes one training iteration (or epoch) of MRCNN. The next epoch again begins with the selection of salient-voxel-zones using 3DGradCAM with the updated parameter $\theta_1$ instead of the pre-trained parameters. The complete process of incorporating the interpretability based feedback system is shown in Figure~\ref{Fig:feedback} and explained in Algorithm~\ref{Alg:Feedback}.
}

\section{Experimental Results \& Discussion}
\label{Sec:Results}

In this section, we present the classification results of the proposed MRCNN framework on ModelNet10 and ModelNet40 datasets~\cite{wu20153d} that contain 3D geometric models of 10 and 40 different categories respectively. The 3D models are voxelized using the voxelization scheme mentioned in Section~\ref{Sec:Voxelization}, yielding a set of coarse voxel grid and fine voxel grids with a single resolution of $8^3$ and $32^3$ respectively. Additionally, we also voxelized two sets of multi-resolution data to test the efficacy of MRCNN; a $8^3$ coarse voxel grid with a $4^3$ fine voxel grid giving an effective resolution of $32^3$ resolution and a $32^3$ coarse voxel grid with a $4^3$ fine voxel grid, resulting in a effective resolution of $128^3$. We conducted a set of experiments on the 4 different resolutions of data and compared the classification performance between a \emph{Coarse-Level CNN} applied on the coarse and dense resolution data and MRCNN applied on the multi-resolution data. For the multi-resolution data, we applied our proposed MRCNN by randomly sampling 40\% of the \emph{coarse-level} boundary voxels, and used the fine resolution voxels of these coarse boundary voxels as input to the \emph{Fine-level CNN}. We then selectively embed the output of \emph{Fine-level CNN} in the coarse level boundary voxels and continue the forward pass. Empirically, we find that sampling 40\% of boundary voxels gives a good classification performance without prolonging the training time excessively.

\begin{figure}[!b]
 \centering
 \includegraphics[width=0.46\textwidth,clip,trim={0.13in 0in 0in 0.1in}]{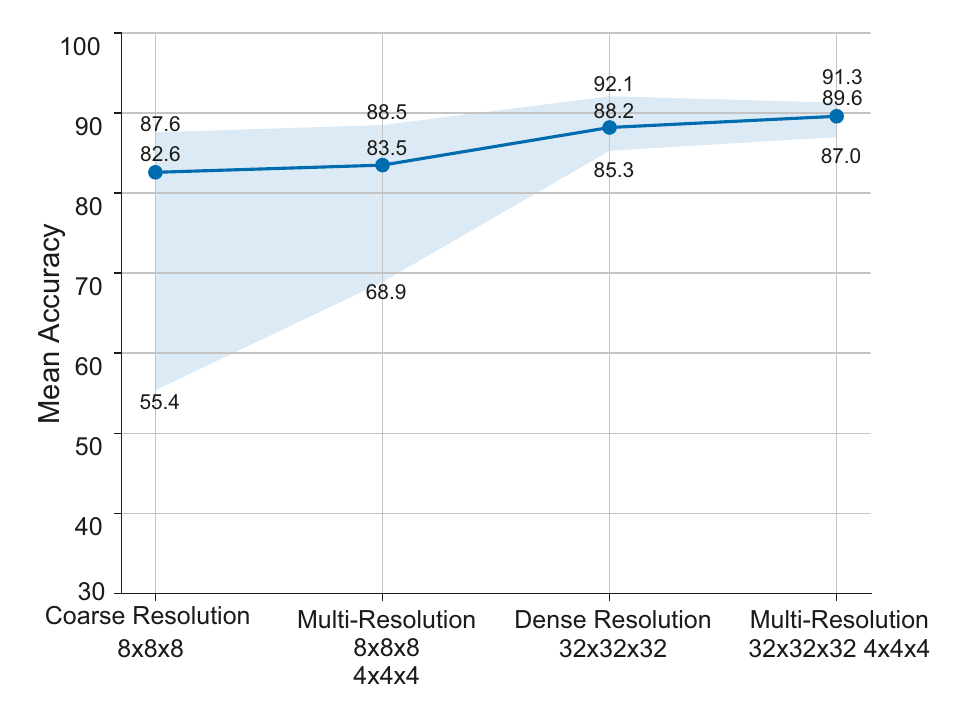}
 \caption{Mean classification accuracies with different input resolutions on ModelNet10 dataset. Coarse and dense resolutions are trained with a conventional 3DCNN while the multi-level voxel grids are trained with MRCNN.}
 \label{Fig:Accuracy} 
\end{figure}

Figure~\ref{Fig:Accuracy} shows the mean test accuracy of object classification using MRCNN on ModelNet10 test dataset by running multiple inferences with various network hyperparameters. Variance in the classification accuracies are represented by the shaded region. We see that there is a clear trend showing better performance for higher effective resolution.
Comparing the performance of a regular CNN on the coarse $8^3$ resolution data with the performance of MRCNN on  multi-resolution data, it is evident that MRCNN enables has better performance. Subsequently, a regular CNN applied on a dense voxel grid of $32^3$ is able to achieve a slightly better classification accuracy than both. Due to memory constraints of GPUs, we are unable to demonstrate the performance of a \emph{Coarse-level CNN} applied on dense resolution data of~$128^3$. Nonetheless, using MRCNN, we are able to train and achieve the best classification performance using an effective resolution of $128^3$ represented by a coarse resolution of $32^3$ and a finer resolution of $4^3$.

\begin{table}[t]
    \setlength\extrarowheight{3pt}
    \small
    \caption{Comparison of deep learning frameworks with voxel based representation for ModelNet10 object recognition. $^*$~represents value interpreted from plot}
    \vspace{-0.15in}
    \begin{center}

    \label{Table:comparison}
      \begin{tabular}{|l |c| c|}
        \hline 
        Method          &   Data Representation &   Accuracy \% \\
        \hline
        \rowcolor{grey}
        \textbf{MRCNN}         &   \textbf{Multi-level voxels}  &   \textbf{91.3}   \\
        \hline
        \rowcolor{grey}
        OctNet~\cite{riegler2016octnet}          &   Octree Voxels       &   $91.0^*$      \\
        \hline
        
        3D Shapenets~\cite{wu20153d}    &   Voxels              &   83.5    \\
        \hline
        VoxNet~\cite{7353481}          &   Voxels              &   92.0      \\
        \hline
        Beam Search~\cite{7900177}     &   Voxels              &   88.0      \\
        \hline
        3DGAN~\cite{Wu:2016:LPL:3157096.3157106}           &   Voxels              &   91.0      \\
        \hline
        binVoxNetPlus~\cite{8403889}   &   Voxels              &   92.3  \\
        \hline
        LightNet~\cite{ZHI2018199}        &   Voxels                  &   93.9      \\
        \hline
      \end{tabular}
    \end{center}
\end{table}

\begin{table}[b!]
    \setlength\extrarowheight{3pt}
    \small
    \caption{Comparison of deep learning frameworks with voxel based representation for ModelNet40 object recognition. $^*$~represents value interpreted from plot.}
    \vspace{-0.15in}
    \begin{center}

    \label{Table:comparison40}
      \begin{tabular}{|l |c| c|}
        \hline 
        Method                  &   Data Representation             &   Accuracy \%         \\
        \hline
        \rowcolor{grey}
        \textbf{MRCNN}          &   \textbf{Multi-level voxels}     &   \textbf{86.2}       \\
        \hline
        \rowcolor{grey}
        OctNet                  &   Octree Voxels                   &   $85.5^*$            \\
        \hline
        
        3D Shapenets            &   Voxels                          &   77.3                \\
        \hline
        VoxNet                  &   Voxels                          &   83.0                \\
        \hline
        Beam Search             &   Voxels                          &   81.26                \\
        \hline
        3DGAN                   &   Voxels                          &   83.3                \\
        \hline
        binVoxNetPlus           &   Voxels                          &   85.47                \\
        \hline
        LightNet                &   Voxels                          &   88.93                \\
        \hline
      \end{tabular}
    \end{center}
\end{table}

\comment{
\begin{figure*}[t!]
 \centering
 \includegraphics[width=0.82\textwidth,clip,trim={0in 0in 1.0in 0.0in}]{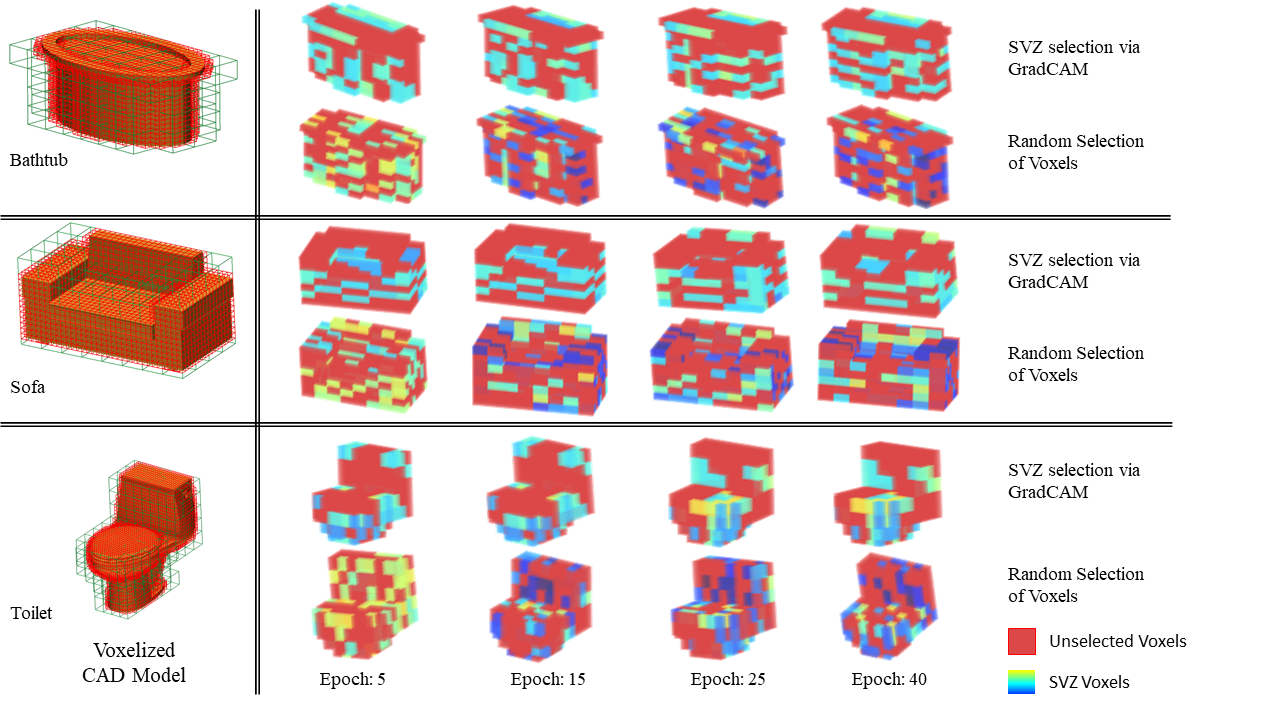}
 \caption{Visualization of salient-voxel-zones from \emph{Fine-level CNN} to \emph{Coarse-level CNN} via random selection vs. GradCAM on ModelNet10 dataset using MRCNN. We can observe that the intensity of the embedding changes drastically between the $5^{th}$ and $15^{th}$ epoch, which demonstrates the learning of a region's importance. Hence, adaptive SVZ selection technique enables a more efficient sampling of important boundary voxels}
 \label{Fig:voxelzone}
\end{figure*}
}

Comparisons of our object classification results with the performance of other spatial deep learning methods are tabulated in Tables~\ref{Table:comparison} and \ref{Table:comparison40} for ModelNet10 and ModelNet40 dataset respectively. We highlight the performance of MRCNN with respect to OctNet due to the similarities in data representation (high resolution voxel grid) and classification task that exploits the sparsity in spatial data in both the frameworks. In addition to that, we compare MRCNN performance with other voxel based methods employed on the ModelNet datasets. We can see that MRCNN (91.3\%) outperforms some of the voxel based methods and is better at classification than OctNet (91.0\%) for ModelNet10. A similar trend is seen in ModelNet40 classification accuracies. 

\begin{figure}[!t]
 \centering
 \includegraphics[width=0.45\textwidth,clip,trim={0in 0in 0.0in 0.5in}]{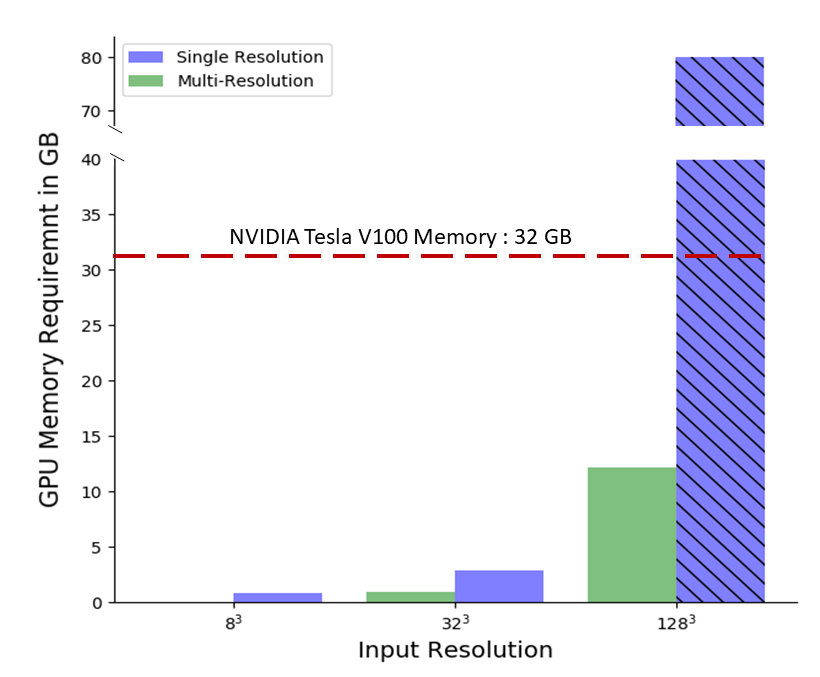}
 \caption{GPU memory usage of MRCNN training \& equivalent CNN training on specified voxel grid resolutions. Red horizontal line shows the current prominent GPU capacity. Blue hatched bar depicts the anticipated memory usage while training a $128^3$ dense voxel grid on CNN.}
 \label{Fig:memory}
\end{figure}

An additional advantage of the MRCNN framework is lower GPU memory utilization during training of the network. In Figure~\ref{Fig:memory}, we show a comparison between the memory requirements of the GPU for training on four different resolutions of voxel data with constant batchsize. The memory required by a GPU scales polynomially ($n^3$) with the voxel grid resolution $n$, hence we were unable to train a dense-level network on $128^3$ voxel resolution (shown as a blue hatched bar). We can see that MRCNN training with multi-level voxel grid representations utilizes considerably less memory than a dense CNN network training on the same effective resolution dense voxel grid.
This highlights the effect of sparsity where the increase in classification performance scales non-linearly with data resolution.

\section{Conclusions}
\label{Sec:Conclusions}

In this paper, we explore a novel deep learning architecture, MRCNN, to learn from 3D data in a hierarchical manner using multi-level voxel-based data structures. 
Our object recognition results show that MRCNN performance is significantly better and robust compared to that of the regular CNNs trained on coarse-resolution data while having similar memory requirements. MRCNN also performs almost as well as CNNs trained on dense data with equivalent resolution while keeping the memory requirements significantly lower. Future works will include exploring efficacies of MRCNN on various object recognition datasets as well as other relevant computer vision problems where extraction of multi-scale features is critically important. 

\section*{Acknowledgements}
This paper is part of the Deep Learning for Geometric Shape Understanding workshop~\cite{SkelNetOn19} in IEEE Computer Society Conference on Computer Vision and Pattern Recognition (CVPR'2019). This work is supported in part by NSF under Grant No. CMMI:1644441.



{\small
\bibliographystyle{plainnat}
\bibliography{arxiv}
}

\end{document}